\documentclass{article}

\usepackage[final]{corl_2018} 

\usepackage{wrapfig}
\usepackage{amsmath}
\DeclareMathOperator*{\argmin}{arg\,min}
\usepackage{graphicx}
\graphicspath{{figures/}}

\title{HybridNet: Integrating Model-based and Data-driven Learning to Predict Evolution of Dynamical Systems} 

\author{
  Yun Long\\
  Georgia Tech \\ 
  \texttt{yunlong@gatech.edu} \\
  \And
  Xueyuan She \\
  Georgia Tech \\
  \texttt{xshe6@gatech.edu} \\
  \AND
  Saibal Mukhopadhyay \\
  Georgia Tech \\
  \texttt{saibal.mukhopadhyay@ece.gatech.edu} \\
}

\begin{document}
\maketitle



\begin{abstract}
The robotic systems continuously interact with complex dynamical systems in the physical world. Reliable predictions of spatiotemporal evolution of these dynamical systems, with limited knowledge of system dynamics, are crucial for autonomous operation. In this paper, we present HybridNet, a framework that integrates data-driven deep learning and model-driven computation to reliably predict spatiotemporal evolution of a dynamical systems even with in-exact knowledge of their parameters. A data-driven deep neural network (DNN) with Convolutional LSTM (ConvLSTM) as the backbone is employed to predict the time-varying evolution of the external forces/perturbations. On the other hand, the model-driven computation is performed using Cellular Neural Network (CeNN), a neuro-inspired algorithm to model dynamical systems defined by coupled partial differential equations (PDEs). CeNN converts the intricate numerical computation into a series of convolution operations, enabling a \textit{trainable} PDE solver. With a feedback control loop, HybridNet can learn the physical parameters governing the system's dynamics in real-time, and accordingly adapt the computation models to enhance prediction accuracy for time-evolving dynamical systems. The experimental results on two dynamical systems, namely, heat convection-diffusion system, and fluid dynamical system, demonstrate that the HybridNet produces higher accuracy than the state-of-the-art deep learning based approach.
\end{abstract}


\section{Introduction}
	
Modeling and prediction of spatiotemporal behavior of complex physical systems is an important problem in science and engineering. The physical systems are mostly defined by coupled partial differential equations (PDEs). Traditionally, computationally expensive numerical methods running on high-performance computing systems have been used to study these systems. The success of deep learning has motivated recent developments in machine learning algorithms for analysis and forecasting of physical systems, for example,  motion tracking \cite{nips2017mt, vp_gan2}, video prediction \cite{nips2017vp, vp_gan1, nips2016vp1}, weather forecasting \cite{convlstm}, to name a few. The deep learning based approaches promise improved speed of prediction, thanks to the extensive research in energy-efficient algorithms and hardware for deep learning \cite{isca,dadiannao}, making it feasible to run real-time forecasting in power-constrained mobile platforms, such as in robotic agents or smart phones.   
However, DNNs are solely data-driven and lacks consideration for the internal system dynamics nor physical mechanism. Any time-dependent variation in system dynamics or parameters (such as velocity, force, pressure, etc) degrades the effectiveness of the purely data-driven approach to modeling of dynamical systems.

\begin{figure}[t]
\includegraphics[trim={4cm 9cm 4cm 6cm},clip, width=0.99\linewidth]{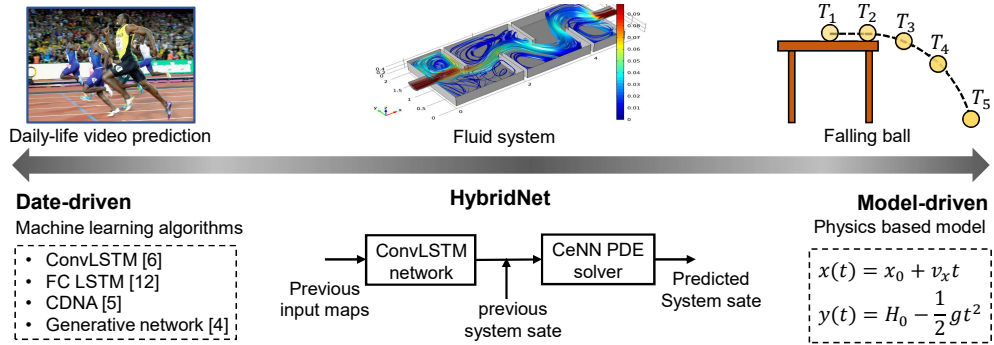}
\caption{An illustration of different modeling approaches for dynamical systems.}
\centering
\label{figure1}
\end{figure}

There exist dynamical systems that are very hard to model with explicit physical equations; DNN-based data-driven predictions is attractive for such systems (left end of Figure \ref{figure1}). Likewise, there are systems that can be fully defined by physical principles, and governing equations for systems, inputs, and all parameter values are known; numerical computations of the physical model is appropriate for such systems (right end of Figure \ref{figure1}). However, most real-world applications exist in between where only in-exact knowledge of system/input dynamics or physical parameters are available. For example, as shown in the middle of Figure \ref{figure1}, we know the dynamics of fluid system is governed by Navier-Stokes equations. However, we can't solve it without knowing the geometry of the system, external forces and other physical parameters such as material density, viscosity, etc. For such problems, we argue that it is important to integrate model-driven computation and data-driven learning. 


This paper presents HybridNet that couples deep learning algorithms and model-driven computation to accurately predict spatiotemporal evolution of dynamical systems. HybridNet consists of two interacting parts: \par
First, at the front-end, Convolutional LSTM (ConvLSTM) \cite{convlstm} is used as the data-driven deep learning algorithm. Unlike the classical LSTM, which performs input-to-state and state-to-state transition with dense connections (i.e. fully connected), ConvLSTM has all the input, hidden state, output and gates as 3-D tensor with uniform spatial dimensions. The internal transition is conducted in a convolutional fashion to retain the spatial information during processing \cite{convlstm}. We utilize ConvLSTM to \textit{predict the evolution of external perturbation/force} (i.e. input) to a system. Examples of perturbation can be the moving heat source/sink in a heat dissipation system, an revolving obstacle in a fluid dynamic system or more realistically, a tropical cyclone on earth. The benefits of using ConvLSTM network to predict the motion pattern of external perturbation are two-fold: first, for most dynamics systems, the perturbation can be easily measured or access (with only a few sensors) than the system state (typically requires measurements for each mesh grid); Second, intuitively, learning the spatiotemporal pattern of the external forces is easier than modeling the dynamics which can be highly non-linear. \par

Second, we use Cellular Neural Network (CeNN) \cite{CeNN}, a neuro-inspired algorithm with highly parallel computation fabric to solve coupled PDEs. 
We show that CeNN transforms numerical computations in a PDE solver to iterative convolution operations and hence, can be efficiently solved using optimized machine learning frameworks (such as Tensorflow and Caffe). Moreover, the convolution based operation in CeNN facilitates \textit{learning unknown physical parameters} (such as diffusion coefficient in heat system or material density in fluid system) using standard back-propagation algorithms without explicit definition of gradients for each physical/mathematical equations. Moreover, with a CeNN-based 'trainable' PDE solver, the system can even adaptively refine the model when system parameters change over time. The details of CeNN based PDE solver are discussed in Section \ref{CeNN}. \par

We evaluate HybridNet with two applications. First, we consider a simple heat dissipation system with moving heat sources. Second, we study fluid dynamic system defined by Naiver-Stokes equations, which is important for many robotic applications such as underwater robot, soft robot, and aero/hydro-dynamics optimization. Our experiments\footnote{All experiments are implemented with Python and Tensorflow running on a NVIDIA GTX 1080 Ti GPU. All the source code is available at \url{https://github.gatech.edu/ylong32}.} show that the proposed method produces more accurate prediction, when compared to results from solely data-driven (machine learning) approach or purely model-driven numerical solver with in-complete knowledge of dynamics/parameters. 


\section{Related Work}
\label{relatedworks}
Recently, machine learning based spatiotemporal processing have achieved higher accuracy than conventional approaches such as optical flow based methods \cite{optical_flow}. In \cite{vphistory1}, a recurrent neural network architecture is utilized for both video frame prediction and language modeling. In \cite{vphistory2}, an LSTM based encoder-decoder architecture is proposed for video reconstructing and predicting. An action conditional auto-encoder model is developed to predict next frames of Atari games in \cite{vphistory3}. To enhance the spatial correlation of the classical LSTM network, ConvLSTM is proposed to deal with the spatiotemporal sequence forecasting problems \cite{convlstm}. More recently, generative adversarial network (GAN) is extensively researched for production of plausible video frames \cite{nips2017vp, vp_gan1, vp_gan2}.\par

Conventionally, modeling and predicting dynamical systems is conducted with numerical computing (i.e. solving multiple coupled PDEs) \cite{weather_numerical, isca}. Recently, along with the success of machine learning, modeling the dynamical system with data-driven approaches have attracted research attention and produces satisfactory results for several scientific problems \cite{convlstm, ibm_tide, michigan, googleCFD_ml, pgnn}. Singh et al., train a neural network to select the best model and parameters for the turbulence modeling task \cite{michigan}. Tompson et al., utilize DNN to accelerate the simulation of Eulerian fluid system \cite{googleCFD_ml}. Recently, Karpatne et al., propose a physical-guided neural network (PGNN) to model the lake temperature which leverages the output of physical system to generate prediction using a multi-layer perceptron \cite{pgnn}. \par

There are also efforts in training robots to learn system dynamics. Guevara et al., propose using approximate fluid simulation to teach robots not to spill \cite{corl2017-1}. Whitman et al., present a differentially-constrained machine learning model to learn physical phenomena for robotic design tasks \cite{corl2017-2}.

Compared to the prior works, this paper makes following unique contributions:
\begin{itemize}
    \item We present a hybrid network that couples data-driven learning to predict external forces (using ConvLSTM) with model-driven computation (with CeNN) for system dynamics. 
    \item We present CeNN with trainable template as a neuro-inspired algorithm for computing the dynamical system model that transforms PDE solution to iterative convolution operations.
    \item We demonstrate that in a CeNN based model-driven computation, templates can be trained with backpropagation algorithm to learn unknown physical parameters.
    \item We develop a feedback-driven algorithm for real-time adaptation of the HybridNet (specifically, the CeNN templates), to enable accurate  forecasting even with systems with time-evolving physical parameters.
\end{itemize}

\section{CeNN as PDE solver}
\label{CeNN}

\begin{figure}[t]
\includegraphics[trim={0.5cm 8cm 1cm 5.5cm},clip, width=0.99\linewidth]{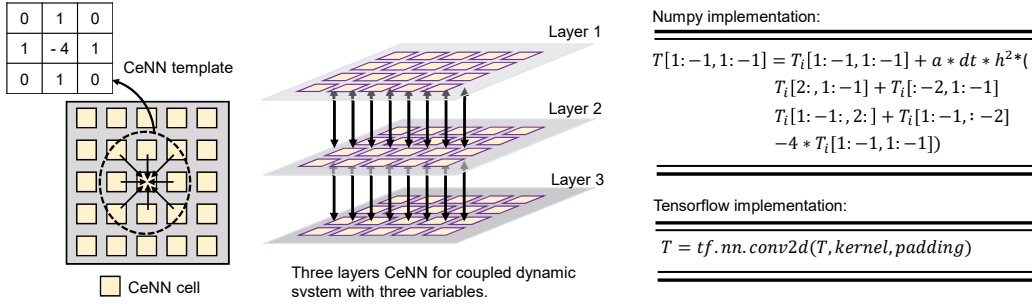}
\caption{(a) A 2-D CeNN array. Cells are locally connected. (b) Multi-layers CeNN with each layer represents one variable to form coupled and complex system dynamics. (c) Pseudo code for heat equation solving with native numpy array implementation and convolution operations.}
\centering
\label{figure2}
\end{figure}

CeNN is a novel algorithm proposed by Chua and Yang \cite{CeNN}. A single layer CeNN is composed of a set of cells organized as a 2-D array, shown in Figure \ref{figure2}(a). Each cell in CeNN follows an ordinary differential equation (ODE). Each cell is connected to a set of neighbouring cells and external inputs using feedback and feedforward templates, respectively. The template weights define dynamics of the system. \par

The behavior of each cell in CeNN is defined by the following equation: 
\begin{equation}
\frac {\partial x_{ij}(t)} {\partial t} = -x_{ij}(t)+\sum_{C(k,l) \in N_{r}(i,j)}  {A}_{kl} \cdot x_{kl}(t) +\sum_{C(k,l) \in N_{r}(i,j)} {B}_{kl} \cdot u_{kl}(t)+z
\end{equation}
where $i$ and $j$ are row and column location index, $x_{ij}(t)$ is the cell state, $u_{kl} (t)$ is the external input, and $z$ is the offset. ${A}_{kl}$ is the cell state interaction template (feedback template) which represent the impact of cell's neighborhood, $B_{kl}$ is the template of input from external source or other layers (feedforward template). Here, $N_{r}(i; j)$ represents the scope of intercommunication region (i.e. connected neighbors). $C(k;l) \in Nr(i; j)$ indicates cell $(k; l)$ is inside the intercommunication region of cell $(i; j)$. 
As shown in Figure \ref{figure2}(b), CeNN with multiple coupled layers can construct more complex system where the dynamics are described by coupled differential equations \cite{isca, CeNN2}.\par

We use heat equation as an example to illustrate how to map PDE onto CeNN. Please refer to the supplementary materials for example mapping of more complex and coupled dynamics. Heat equation and its discretized form are given by:
\begin{equation}
\frac{\partial T(x,y,t)}{\partial t} = K \cdot \Delta T(x,y,t)
\end{equation}
\begin{equation}
\frac{\partial T(x,y)}{\partial t} = K \cdot \{ \frac{T(x+h,y)+T(x-h,y)-2  T(x,y)}{h^{2}} + \frac{T(x,y+h)+T(x,y-h)-2  T(x,y)}{h^{2}}\}
\end{equation}
where $T(x,y,t)$ is the temperature at location $(x,y)$ and time t, $K$ is the heat diffusion coefficient and $\Delta$ is the Laplace operator (equals to ${\nabla}^{2}$ or $\nabla \cdot \nabla$). $h$ is the step size in 2-D Euclidean space. Equation (1) and (3) are essentially identical if we define the CeNN templates as follow:
\[
A= K \cdot
  \begin{bmatrix}
    0       & 1/h^{2}    & 0 \\
    1/h^{2} & -4/h^{2}+1 & 1/h^{2} \\
    0       & 1/h^{2}    & 0
  \end{bmatrix},
B=0,
Z=0.
\]

We observe that CeNN provides a unique approach to build a general purpose, trainable PDE solver by converting PDEs to convolution operations. For example, mapping heat equation to CeNN gives a space-invariant template (all cells in the CeNN share the same templates: $A, B$ and $Z$). Therefore, the cell state ($x_{ij}$), which represents temperature at each grid, can be updated with convolutional operations. To be more specific, the 2-D heat map recording the temperature at each spatial grid can be treated as an input feature map of a convoluational layer with input channel size 1; the template ($A$) is then used as a $3 \times 3 \times 1$ kernel to perform convolution operation over the feature map. The pseudo code in Figure \ref{figure2} illustrates the classical array based implementation as well as our convolution based method (using Tensorflow) for solving heat equation. In summary, with CeNN based PDE solver, we perform the numerical computing in a machine learning fashion (i.e. using the convoluational layer), keeping the gradients for back-propagation and making the numerical computing also 'trainable'. \par

\section{The proposed model: HybridNet}
\label{model}

\begin{figure}[t]
\includegraphics[trim={0cm 6.5cm 0cm 7cm},clip, width=0.99\linewidth]{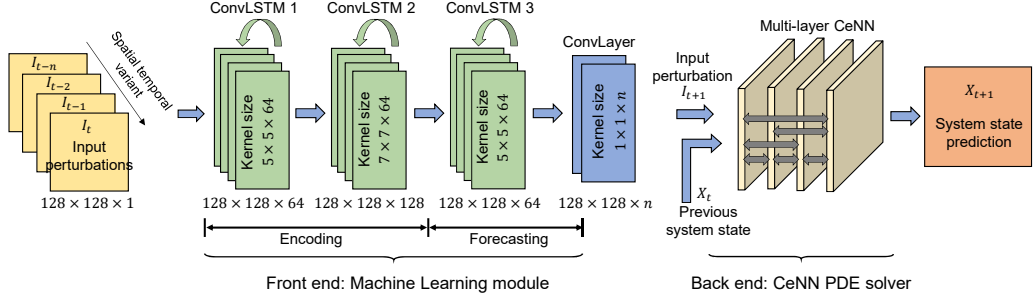}
\caption{Architecture of HybridNet including multi-layer ConvLSTM network to learn and predict the external perturbation, and CeNN based PDE solver to learn the unknown coefficients and perform computation for system dynamics.}
\centering
\label{figure3}
\end{figure}

Figure \ref{figure3} shows the architecture of HybridNet. The front end composes of multiple stacked ConvLSTM, receiving a series of input maps recording the past information of external forcing/perturbation. The output of ConvLSTM network is the prediction of perturbation map for the next time step. For example, considering there is an moving obstacle inside a fluid system, the ConvLSTM network will predict the location of the obstacle based on its previous locations. \par

The front end ConvLSTM network can be further divided into two parts: encoding and forecasting. The encoding network contains two stacked ConvLSTM with 64 and 128 output channels, respectively. The input/state as well as convolution kernel size are annotated in Figure \ref{figure3}. The forecasting network consists of one ConvLSTM with 64 output channels and a convolutional layer to squeeze the output from the last ConvLSTM into a 3-D tensor as the predicted input map at $T_{n+1}$. It should be noted that the third dimension of this tensor is application-dependent, equals to the number of variables inside the input map. We explored increasing the size of the encoding-forecasting network via adding more ConvLSTM layers, downsampleing/upsampling the feature map size, and integrating skip connections. We observed that such modifications deliver trivial accuracy improvements for the tested application while slowing down the training/inference speed. \par

At the back end, the CeNN takes the output from ConvLSTM networks as input, perform model-driven computation (solving PDEs) and output the system state (e.g. a temperature map in heat system) for the next time step. Since our model predicts one frame per cycle, we then roll-out the model, passing in the prediction from the previous time step to generate new prediction.\par

The size of each CeNN layer is identical to the ConvLSTM network. The number of CeNN layers and templates is also application-dependent. For example, in heat diffusion and convection system, there are one layer and two templates  (for diffusion and convection, respectively). In the Navier-Stokes system, there are 5 layers and 13 templates (depending on the physical principles, layers are coupled together with templates). Moreover, since we perform the numerical computing with time discretization, an internal while loop is employed to perform the convolution iteratively.

\section{Training and Real-time Learning in HybridNet}


\textbf{Train the ConvLSTM for perturbation prediction}:
The objective of our ConvLSTM network is to predict the perturbation map at the next time step ($I_{t+1}$) based on the observation of a sequence of previous perturbation maps ($I_{t}, I_{t-1}, I_{t-2}, ... I_{t-N}$) by minimizing the prediction loss function\footnote{The loss function combines of L1-norm and L2-norm: $Loss=\alpha \sum_{i,j} |{\hat Y}_{ij}-Y_{ij}| + \beta \sum_{i,j} ({\hat Y}_{ij}-Y_{ij})^{2}$ where $\alpha=0.2$, $\beta=0.8$}:
\begin{equation}
\argmin_{f_{ConvLSTM}} Loss({\hat I_{t+1}}, I_{t+1}) \quad where \quad {\hat I}_{t+1}=f_{ConvLSTM}(I_{t}, I_{t-1}, I_{t-2}, ... I_{t-N})
\end{equation}
During training, we take 5 frames of perturbation maps as known information to predict the perturbation map at next time step. We employ Adam optimizer (with initial learning rate = 0.001) for the training since it results in better convergence than RMSProp and SGD in our experiments.


\textbf{Learn physical parameters with CeNN}:

Thanks to the trainability, CeNN based PDE solver has the ability to recognize the unknown physical parameters by minimizing the mismatch between the computed system state ${\hat X}_{t+1}$ and the ground truth $X_{t+1}$.
\begin{equation}
\argmin_{f_{CeNN}} Loss({\hat X_{t+1}}, X_{t+1}) \quad where \quad {\hat X}_{t+1}=f_{CeNN}({\hat I}_{t+1}, X_{t})
\end{equation}
In our training approach, first, the CeNN is programmed to map the system dynamics (i.e. PDEs) by defining the coupling between nodes and layers. The template weights related to the unknown physical parameters (such as heat diffusion coefficient) are kept trainable. Next, the training starts with random initialization of the physical parameters. Standard back-propagation algorithm using Stochastic Gradient Descent (SGD) is utilized for the coefficient regression. This is similar to the training of ConvLSTM except that CeNN training uses much larger initial learning rate and decay rate since there are only a few parameters are trainable inside CeNN. The trainable CeNN allows us to learn the parameters of a specific system from data, rather that depending on exact knowledge of the parameter values.

\begin{wrapfigure}{R}{0.5\textwidth}
  \vspace{-4mm}
  \begin{center}
    \includegraphics[width=0.48\textwidth]{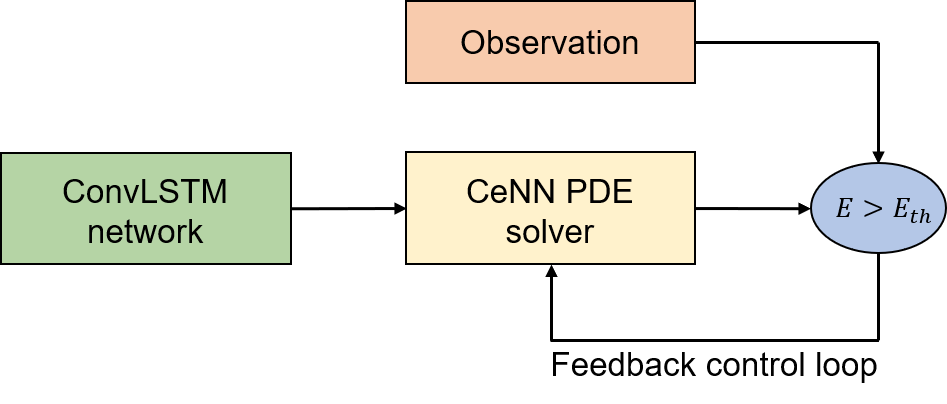}
  \end{center}
  \vspace{-4mm}
  \caption{CeNN can adaptively re-learn the coefficient via a feedback control loop.}
  \vspace{-4mm}
  \label{figure4}
\end{wrapfigure}

The more intriguing opportunity of having a trainable PDE solver is the feasibility of real-time learning of the parameters using a feedback control loop. This is very useful considering that parameters of real-world physical systems are often not fixed and can change over time. Our approach is shown in Figure \ref{figure4} that considers availability of observed data (for example, measurements from sensors or cameras). Once the mismatch between the observation and predicted output from CeNN becomes larger than the preset threshold, the CeNN is informed by the feedback control loop to re-learn the coefficient. This ensures the system can always provide accurate prediction even the system parameters change over time. This is essentially a reinforcement learning system where the robots (i.e. agents) interact with the physical systems (i.e. environments) and minimize the prediction loss (maximize the rewards) based on observation rather than training data. Moreover, the adaptive re-learning feature also provides a new approach to approximate some physical parameters especially when these values are difficult to measure or can't be derived from math equations. Should be noted that we only consider the error caused by coefficients changing rather than the change of perturbation pattern nor the physical laws. Therefore, only the coefficients is trainable and all other variables inside the network are frozen when re-learn the coefficients.

\section{Experimental Results} 


\subsection{Heat diffusion and convection system}

For heat convection-diffusion system, the system size is $128 \times 128$. We consider using a moving heat source to represent the perturbation. The heat source is a round region with radius equals to 20. The initial location of heat source is randomly selected. The moving direction as well as the moving velocity of the heat source are also randomly chosen but once initialized, stay fixed. Then, we calculate the system state (temperature at each grid) numerically based on the locations of heat source following two types of dynamics: heat convection and diffusion\footnote{The function for heat dissipation and convection is: $\frac{\partial T_{xy}}{\partial t} = C \cdot (T_{external} - T_{xy}) + K \cdot \Delta T_{xy}$, $C$ and $K$ are convection and diffusion coefficient, respectively.}.\par

\begin{figure}[t]
\includegraphics[trim={0.5cm 7.5cm 1cm 6cm},clip, width=1.0\linewidth]{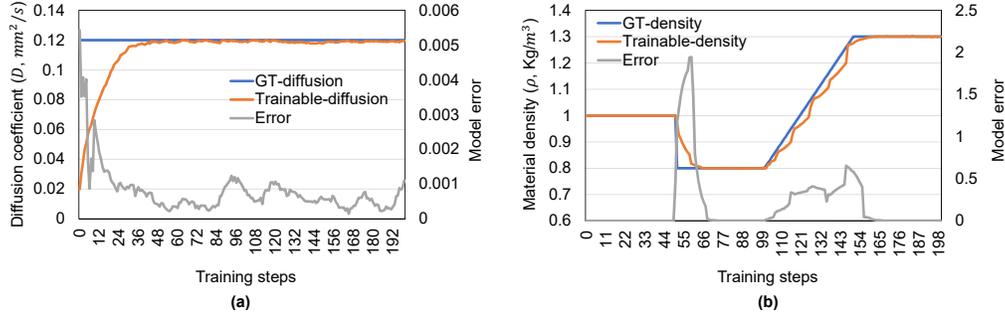}
\caption{(a): Heat dissipation system: Learn the diffusion coefficient from a random value. (b) Fluid system: Adaptively learn the material density. Prediction error is shown in grey line.}
\centering
\label{figure7}
\end{figure}

\textbf{Learning heat diffusion coefficient with CeNN}: 
As shown in Figure \ref{figure7}(a), we randomly initialize the heat diffusion coefficient. The error is large at the beginning but quickly drops, meanwhile the value of diffusion coefficient converges to the ground truth. 

\begin{figure}[t]
\includegraphics[trim={1cm 6cm 1.5cm 5.5cm},clip, width=0.99\linewidth]{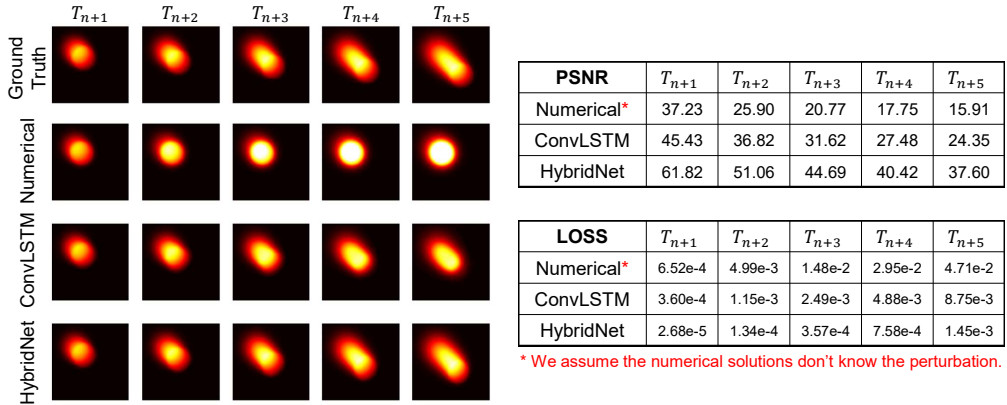}
\caption{(a-c): Qualitative and quantitative comparison between HybridNet,  numerical approach (without knowing the perturbations) and machine learning approach. Models predict 5 frames of system state. PSRN and LOSS are used to evaluate the prediction accuracy.}
\centering
\label{figure5}
\end{figure}

\textbf{Forecast System Evolution with HybridNet}: 
We now demonstrate forecasting performance of HybridNet with learned physical parameters. We compare HybridNet with both numerical method and machine learning method. The numerical approach solves heat equation without knowing the heat source motion. The machine learning approach utilizes ConvLSTM network (the front end of HybridNet) solely to predict the heat map (essentially, it can be viewed as a classical video prediction network similar with the one proposed in \cite{convlstm}). 
Figure \ref{figure5}(a) demonstrates the ground truth and predicted heat maps from different configurations. We also quantitatively evaluate the accuracy of different configurations based on Peak Signal to Noise Ratio (PSNR) \cite{vp_gan1} and our own LOSS function (shown in the table inside Figure \ref{figure7}). Note that for PSNR, larger value indicates a smaller mismatch while for LOSS larger value indicates a larger mismatch. HybridNet consistently outperforms other configurations even though all methods tend to have a lower accuracy for long-term prediction. \par

\subsection{Navier-Stokes equations for fluid dynamics systems}

\begin{figure}[t]
\includegraphics[trim={1.1cm 6cm 1.6cm 5.8cm},clip, width=1.0\linewidth]{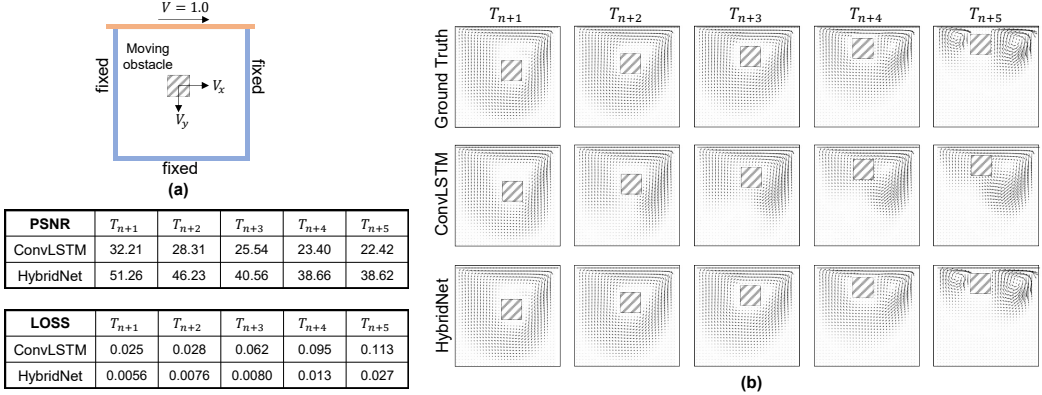}
\caption{(a): 2-D fluid system with a driven lid and a moving obstacle. (b) Visualization of flow velocity for ground truth, prediction of ConvLSTM network, prediction from HybridNet. Please refer to \url{https://github.gatech.edu/ylong32} for high resolution figures. Insert table: quantitatively analyses for prediction accuracy in terms of PSNR and LOSS.}
\centering
\label{figure6}
\end{figure}

The dynamics of fluid system is governed by Navier-Stokes equations\footnote{Navier-Stokes equations: $\frac{\partial\Vec{v}}{\partial t}+(\Vec{v} \cdot \nabla)\Vec{v}=-\frac{1}{\rho}\nabla p + \nu {\nabla}^{2} \Vec{v}$ and $\nabla \cdot \Vec{v}=0$, representing momentum and mass conservation, respectively. The original CFD implementation can refer to \url{https://github.com/barbagroup/CFDPython}}. Different with linear PDEs in heat diffusion-convection system, the Navier-Stokes equations comprise two coupled nonlinear PDEs, making the system dynamics more complex and unpredictable. We consider a 2-D fluid dynamics system with a driven lid. As shown in Figure \ref{figure6}(a), the top boundary is moving with a fixed speed while other boundaries keep still. An square obstacle (with random initial location and moving direction) is placed inside to disturb the flow pattern. In this work, we only concern the steady state, i.e. we predict the velocity and pressure when the system converges to stationary. Modeling transient sate and turbulence is our next step work.

\textbf{Real-time Learning of Physical Parameters}: 
Rather than learning the physical coefficient from scratch, We evaluate the adaptive learning strategy considering fluid system with changing material density (Figure \ref{figure7}(b)). At the beginning, the density coefficient matches the ground truth value and the error is negligible. Then the density changes abruptly from 1.0 to 0.8 (can be interpreted as a change from water to oil). A large error is detected at once and CeNN start to re-learn the parameters. We also consider the case that the coefficient changes gradually. For example, we gradually inject a new fluid and discharge the original one. Thus, the system contains a mixture of two materials and the density is changing gradually. We observe density coefficient of CeNN can tightly follows the numerical value. Further, our experiments indicate the adaptive re-learning typically only takes a few steps (several seconds of running time on GPU) to converge to the correct value, enabling a real-time self correction system. We argue this is a critical feature for robot design when the robot works at complex, time-evolving dynamical system.

\textbf{Forecast fluid system with HybridNet}: 
As a highly non-linear system, a marginal alternation of perturbation might thoroughly change the system state. For example, as shown in the first row of Figure \ref{figure6}(b), a subtle change of obstacle from time $T_{n+4}$ to $T_{n+5}$ causes very different flow velocity patterns (2 vortices formed at $T_{n+5}$). We train the ConvLSTM network to predict the obstacle motion pattern and CeNN to learn the physical parameters.  We observe that HybridNet can successfully capture this non-linearity, thanks to the CeNN PDE solver which performs numerical computing precisely. On the other hand, the machine learning approach (ConvLSTM network) failed to learn such complex non-linear data representation. We also quantitatively evaluate the predicting accuracy (flow velocity and pressure at each mesh grid) in terms of PSNR and LOSS. Again, HybridNet consistently outperform the machine learning solution.

\subsection{Computational Performance}

We investigate the computational performance of HybridNet considering GPU (measured with NVIDIA GTX 1080Ti) as well as embedded hardware platform using hardware accelerators (estimated) that can be integrated in robotic devices. For ConvLSTM, we project the run-time based on the number reported by DaDianNao \cite{dadiannao}, a well-known DNN accelerator with 20.1W power consumption; for CeNN, we estimate running time with a recent CeNN ASIC accelerator design (1.56W) \cite{isca}. As demonstrated in Table \ref{Hardware}, with dedicated ASICs, the HybridNet can run simulation more than 10x faster than GPU with much lower power budget.  

\begin{table}[t]
\centering
\caption{Running speed of HybridNet on GPU and ASICs}
\begin{tabular}{|c|c|c|c|c|c|c|}
\hline
Running time/step* & \multicolumn{3}{c|}{Running speed @ GPU} & \multicolumn{3}{c|}{Running speed @ ASICs} \\ \hline
                   & ConvLSTM     & \quad CeNN \quad        & Total      & ConvLSTM     & CeNN         & Total       \\ \hline
Heat system        & 0.048 s       & 0.28 s      & 0.33 s     & 2.3 ms        & 3.6 ms       & 5.9 ms      \\ \hline
Fluid system       & 0.051 s       & 2.98 s      & 3.03 s     & 2.5 ms        & 38.7 ms      & 41.2 ms     \\ \hline
\end{tabular}
\raggedright {\small *For heat system and fluid system, each step represent 50 ms and 100 ms real time, respectively.}
\label{Hardware}
\end{table}

\section{Conclusion}

The HybridNet demonstrates the feasibility of integrating data-driven learning and model-driven computation to predict spatiotemporal evolution of dynamical systems. With HybridNet, autonomous agents can forecast system outputs even with in-exact knowledge of input perturbation and can learn physical parameters in a real-time fashion, thereby, enabling higher flexibility when interacting with complex and time-evolving dynamical systems.



\clearpage


\bibliography{example}  

\begin{thebibliography}{21}
\providecommand{\natexlab}[1]{#1}
\providecommand{\url}[1]{\texttt{#1}}
\expandafter\ifx\csname urlstyle\endcsname\relax
  \providecommand{\doi}[1]{doi: #1}\else
  \providecommand{\doi}{doi: \begingroup \urlstyle{rm}\Url}\fi

\bibitem[Jin et~al.(2017)Jin, Xiao, Shen, Yang, Lin, Chen, Jie, Feng, and
  Yan]{nips2017mt}
X.~Jin, H.~Xiao, X.~Shen, J.~Yang, Z.~Lin, Y.~Chen, Z.~Jie, J.~Feng, and
  S.~Yan.
\newblock Predicting scene parsing and motion dynamics in the future.
\newblock In \emph{Advances in Neural Information Processing Systems}, pages
  6918--6927, 2017.

\bibitem[Vondrick et~al.(2016)Vondrick, Pirsiavash, and Torralba]{vp_gan2}
C.~Vondrick, H.~Pirsiavash, and A.~Torralba.
\newblock Generating videos with scene dynamics.
\newblock In \emph{Advances In Neural Information Processing Systems}, pages
  613--621, 2016.

\bibitem[Bhattacharjee and Das(2017)]{nips2017vp}
P.~Bhattacharjee and S.~Das.
\newblock Temporal coherency based criteria for predicting video frames using
  deep multi-stage generative adversarial networks.
\newblock In \emph{Advances in Neural Information Processing Systems}, pages
  4271--4280, 2017.

\bibitem[Mathieu et~al.(2015)Mathieu, Couprie, and LeCun]{vp_gan1}
M.~Mathieu, C.~Couprie, and Y.~LeCun.
\newblock Deep multi-scale video prediction beyond mean square error.
\newblock \emph{arXiv preprint arXiv:1511.05440}, 2015.

\bibitem[Finn et~al.(2016)Finn, Goodfellow, and Levine]{nips2016vp1}
C.~Finn, I.~Goodfellow, and S.~Levine.
\newblock Unsupervised learning for physical interaction through video
  prediction.
\newblock In \emph{Advances in neural information processing systems}, pages
  64--72, 2016.

\bibitem[Xingjian et~al.(2015)Xingjian, Chen, Wang, Yeung, Wong, and
  Woo]{convlstm}
S.~Xingjian, Z.~Chen, H.~Wang, D.-Y. Yeung, W.-K. Wong, and W.-c. Woo.
\newblock Convolutional lstm network: A machine learning approach for
  precipitation nowcasting.
\newblock In \emph{Advances in neural information processing systems}, pages
  802--810, 2015.

\bibitem[Kung et~al.(2017)Kung, Long, Kim, and Mukhopadhyay]{isca}
J.~Kung, Y.~Long, D.~Kim, and S.~Mukhopadhyay.
\newblock A programmable hardware accelerator for simulating dynamical systems.
\newblock In \emph{Proceedings of the 44th Annual International Symposium on
  Computer Architecture}, pages 403--415. ACM, 2017.

\bibitem[Chen et~al.(2014)Chen, Luo, Liu, Zhang, He, Wang, Li, Chen, Xu, Sun,
  et~al.]{dadiannao}
Y.~Chen, T.~Luo, S.~Liu, S.~Zhang, L.~He, J.~Wang, L.~Li, T.~Chen, Z.~Xu,
  N.~Sun, et~al.
\newblock Dadiannao: A machine-learning supercomputer.
\newblock In \emph{Proceedings of the 47th Annual IEEE/ACM International
  Symposium on Microarchitecture}, pages 609--622. IEEE Computer Society, 2014.

\bibitem[Chua and Yang(1988)]{CeNN}
L.~O. Chua and L.~Yang.
\newblock Cellular neural networks: theory.
\newblock \emph{IEEE Transactions on circuits and systems}, 35\penalty0
  (10):\penalty0 1273--1290, 1988.

\bibitem[Horn and Schunck(1981)]{optical_flow}
B.~K. Horn and B.~G. Schunck.
\newblock Determining optical flow.
\newblock \emph{Artificial intelligence}, 17\penalty0 (1-3):\penalty0 185--203,
  1981.

\bibitem[Ranzato et~al.(2014)Ranzato, Szlam, Bruna, Mathieu, Collobert, and
  Chopra]{vphistory1}
M.~Ranzato, A.~Szlam, J.~Bruna, M.~Mathieu, R.~Collobert, and S.~Chopra.
\newblock Video (language) modeling: a baseline for generative models of
  natural videos.
\newblock \emph{arXiv preprint arXiv:1412.6604}, 2014.

\bibitem[Srivastava et~al.(2015)Srivastava, Mansimov, and
  Salakhudinov]{vphistory2}
N.~Srivastava, E.~Mansimov, and R.~Salakhudinov.
\newblock Unsupervised learning of video representations using lstms.
\newblock In \emph{International conference on machine learning}, pages
  843--852, 2015.

\bibitem[Oh et~al.(2015)Oh, Guo, Lee, Lewis, and Singh]{vphistory3}
J.~Oh, X.~Guo, H.~Lee, R.~L. Lewis, and S.~Singh.
\newblock Action-conditional video prediction using deep networks in atari
  games.
\newblock In \emph{Advances in Neural Information Processing Systems}, pages
  2863--2871, 2015.

\bibitem[Richardson(2007)]{weather_numerical}
L.~F. Richardson.
\newblock \emph{Weather prediction by numerical process}.
\newblock Cambridge University Press, 2007.

\bibitem[James et~al.(2017)James, Zhang, and O'Donncha]{ibm_tide}
S.~C. James, Y.~Zhang, and F.~O'Donncha.
\newblock A machine learning framework to forecast wave conditions.
\newblock \emph{arXiv preprint arXiv:1709.08725}, 2017.

\bibitem[Singh et~al.(2017)Singh, Medida, and Duraisamy]{michigan}
A.~P. Singh, S.~Medida, and K.~Duraisamy.
\newblock Machine-learning-augmented predictive modeling of turbulent separated
  flows over airfoils.
\newblock \emph{AIAA Journal}, pages 1--13, 2017.

\bibitem[Tompson et~al.(2016)Tompson, Schlachter, Sprechmann, and
  Perlin]{googleCFD_ml}
J.~Tompson, K.~Schlachter, P.~Sprechmann, and K.~Perlin.
\newblock Accelerating eulerian fluid simulation with convolutional networks.
\newblock \emph{arXiv preprint arXiv:1607.03597}, 2016.

\bibitem[Karpatne et~al.(2017)Karpatne, Watkins, Read, and Kumar]{pgnn}
A.~Karpatne, W.~Watkins, J.~Read, and V.~Kumar.
\newblock Physics-guided neural networks (pgnn): An application in lake
  temperature modeling.
\newblock \emph{arXiv preprint arXiv:1710.11431}, 2017.

\bibitem[Guevara et~al.()Guevara, Taylor, Gutmann, Ramamoorthy, and
  Subr]{corl2017-1}
T.~L. Guevara, N.~K. Taylor, M.~U. Gutmann, S.~Ramamoorthy, and K.~Subr.
\newblock Adaptable pouring: Teaching robots not to spill using fast but
  approximate fluid simulation.

\bibitem[Whitman and Chowdhary(2017)]{corl2017-2}
J.~Whitman and G.~Chowdhary.
\newblock Learning dynamics across similar spatiotemporally-evolving physical
  systems.
\newblock In \emph{Conference on Robot Learning}, pages 472--481, 2017.

\bibitem[Kozek and Roska(1996)]{CeNN2}
T.~Kozek and T.~Roska.
\newblock A double time—scale cnn for solving two-dimensional navier—stokes
  equations.
\newblock \emph{International Journal of Circuit Theory and Applications},
  24\penalty0 (1):\penalty0 49--55, 1996.

\end{thebibliography}

\end{document}